\pdfoutput=1

\documentclass[11pt]{article}

\usepackage{acl}

\usepackage{times}
\usepackage{latexsym}

\usepackage[T1]{fontenc}

\usepackage[utf8]{inputenc}

\usepackage{microtype}

\usepackage{graphicx}
\usepackage{amsmath}
\usepackage{amsthm}
\usepackage{booktabs}
\usepackage{algorithm}
\usepackage{algorithmic}
\usepackage{latexsym}
\usepackage[utf8]{inputenc}
\usepackage{multirow}
\usepackage{pifont}
\usepackage{bm}

\title{Perform Like an Engine: A Closed-Loop Neural-Symbolic Learning Framework for Knowledge Graph Inference}

\author{
Guanglin Niu\textsuperscript{\rm 1},
Bo Li\textsuperscript{\rm 1,2}\thanks{\ \ Corresponding author.},
Yongfei Zhang\textsuperscript{\rm 3,4},
Shiliang Pu\textsuperscript{\rm 5}
\\ 
\textsuperscript{\rm 1} Institute of Artificial Intelligence, Beihang University, Beijing, China\\
\textsuperscript{\rm 2} Hangzhou Innovation Institute, Beihang University, Hangzhou, China\\
\textsuperscript{\rm 3} Beijing Key Laboratory of Digital Media, Beihang University, Beijing, China\\
\textsuperscript{\rm 4} State Key Laboratory of Virtual Reality Technology and Systems, Beihang University,\\Beijing, China
\textsuperscript{\rm 5} Hikvision Research Institute, Hangzhou, China\\
\{beihangngl, boli, yfzhang\}@buaa.edu.cn, pushiliang.hri@hikvision.com
}

\begin{document}
\maketitle
\begin{abstract}
Knowledge graph (KG) inference aims to address the natural incompleteness of KGs, including rule learning-based and KG embedding (KGE) models. However, the rule learning-based models suffer from low efficiency and generalization while KGE models lack interpretability. To address these challenges, we propose a novel and effective closed-loop neural-symbolic learning framework \textbf{EngineKG} via incorporating our developed KGE and rule learning modules. KGE module exploits symbolic rules and paths to enhance the semantic association between entities and relations for improving KG embeddings and interpretability. A novel rule pruning mechanism is proposed in the rule learning module by leveraging paths as initial candidate rules and employing KG embeddings together with concepts for extracting more high-quality rules. Experimental results on four real-world datasets show that our model outperforms the relevant baselines on link prediction tasks, demonstrating the superiority of our KG inference model in a neural-symbolic learning fashion.
\end{abstract}

\section{Introduction}

Typical knowledge graphs (KGs) store triple facts and some of them also contain concepts of entities~\cite{BGF:Freebase}. The KGs have proven to be incredibly effective for a variety of applications such as dialogue system~\cite{dialogue} and question answering~\cite{kbqa}. However, the existing KGs are always incomplete which restricts the performance of knowledge-based applications. Thus, KG inference plays a vital role in completing KGs for better applications of KGs.

\begin{figure}
  \centering
  \includegraphics[scale=0.56]{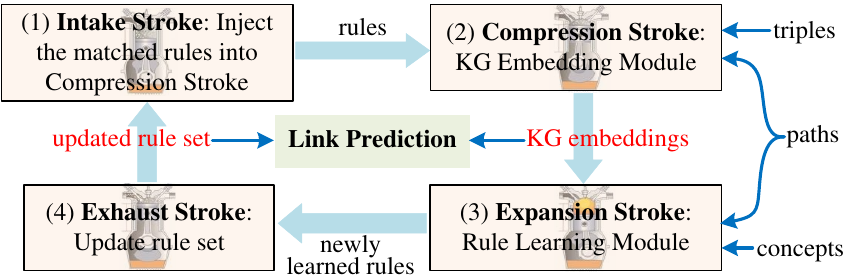}
  \caption{The brief architecture of our closed-loop framework for KG inference EngineKG that performs like a four-stroke engine.}
  \label{fig1}
\end{figure}

The existing KG inference approaches are usually classified into two main categories: (1) Rule learning-based models such as AMIE+~\cite{Galarrage:AMIE} and AnyBurl~\cite{AnyBurl} mine rules from KGs and employ these rules to predict new triples by deduction. However, rule learning-based models suffer from \textbf{low efficiency} of the rule mining process and the \textbf{poor generalization} caused by the limited coverage of inference patterns. (2) KGE technique learns the embeddings of entities and relations to predict the missing triples via scoring each triple candidate, including TransE~\cite{Bordes:TransE}, HAKE~\cite{HAKE} and DualE~\cite{DualE}. The previous KGE models perform in a data-driven fashion, contributing to good efficiency and generalization but \textbf{lacking interpretability}.

Some recent researches attempt to combine the advantages of rule learning-based and KGE-based models to complement each other in a \textbf{neural-symbolic learning fashion}. An idea is to introduce logic rules into KGE models, such as RUGE~\cite{Shu-Guo:RUGE} and its advanced model IterE~\cite{IterE}. These approaches all convert the rules into formulas by t-norm based fuzzy logic to obtain newly labeled triples. However, these models \textbf{cannot maintain the vital interpretability from symbolic rules}. On the other hand, some rule learning-based models succeed in leveraging KG embeddings to extract rules via numerical calculation rather than discrete graph search, including RNNlogic~\cite{RNNLogic}, RLvLR~\cite{RLvLR}, DRUM~\cite{DRUM} and RuLES~\cite{RuLES}. Although the efficiency of mining rules is improved, \textbf{the performance especially generalization of purely employing rules to implement KG inference is still limited}.

To address the above challenges, we propose a closed-loop neural-symbolic learning framework \textbf{EngineKG} via combining an embedding-based rule learning and a rule-enhanced KGE, in which paths and concepts are utilized. Our model is named \textbf{EngineKG} because it performs like an engine as shown in Figure \ref{fig1}:
(1)	\textbf{Intake Stroke}. The closed-path rules (or named chain rules) are injected into the KGE module (analogous to intake) to guide the procedure of learning KG embeddings, where the initial seed rules are mined by any rule learning tool, and the rule set would grow via our designed rule learning module from the first iteration. (2) \textbf{Compression Stroke}. The KGE module leverages the rules and paths to learn the low-dimensional embeddings (analogous to compression) of entities and relations, improving the interpretability and accuracy. (3) \textbf{Expansion Stroke}. The novel rule learning module outputs newly learned rules (analogous to exhaust) by the effective rule pruning strategy based on paths, relation embeddings and concepts.
(4)	\textbf{Exhaust Stroke}. Update the rule set (analogous to exhaust) by merging the previous rule set and the newly learned rules for boosting KGE and KG inference in the next iteration.

Our research makes three contributions:

\begin{itemize}
     \item We propose a novel and effective closed-loop neural-symbolic learning framework that performs embedding-based rule learning and rule-enhanced KGE iteratively, balancing good accuracy, interpretability and efficiency.
    
    \item Paths and ontological concepts are well exploited for supplementing the valuable semantics to both KGE and rule learning, facilitating the better performance of KG inference.
    
    \item The link prediction results and the effectiveness of rule learning on four datasets illustrate that our model outperforms various state-of-the-art KG inference approaches.

\end{itemize}

\section{Related Work}
\label{sec2}

\subsection{Rule Learning-Based Models}
According to the symbolic characteristics of KG, some rule learning techniques specific to KGs are applied to KG inference with relatively good accuracy and interpretability, including AMIE+~\cite{Galarrage:AMIE}, Anyburl~\cite{AnyBurl}, DRUM~\cite{DRUM}, RLvLR~\cite{RLvLR} and RNNLogic~\cite{RNNLogic}. AMIE+ \cite{Galarrage:AMIE} introduces optimized query writing techniques into traditional inductive logic programming algorithms to generate horn rules efficiently. Anyburl learns closed-path rules from KGs in a reinforcement learning framework. DRUM, RLvLR and RNNLogic employ KG embeddings for enhancing efficiency and scalability of rule learning. Whereas, all the previous rule learning algorithms lack generalization since the number of rules mined at one time is very limited.

\subsection{KG Embedding Models}
The typical KG embedding (KGE) models learn the embeddings of entities and relations to measure the plausibility of each triple. TransE~\cite{Bordes:TransE} regards the relations as translation operations from head to tail entities. ComplEx~\cite{Trouillon:ComplEx} embeds the KG into a complex space while DualE~\cite{DualE} embeds relations into the quaternion space to model the symmetric and anti-symmetric relations. HAKE~\cite{HAKE} embeds entities into the polar coordinate system and is able to model the semantic hierarchies of KGs. RUGE~\cite{Shu-Guo:RUGE} and IterE~\cite{IterE} both convert rules into formulas by t-norm fuzzy logic to infer newly labeled triples. Particularly, IterE iteratively conducts rule learning and KG embedding, but \textbf{the significant distinctions between our model EngineKG and IterE} include: (1) \textbf{Usage of rules}: our model leverages rules to compose paths for learning KG embeddings while IterE uses rules to produce labeled triples. Meanwhile, we maintain the interpretability of symbolic rules, while IterE does not. (2) \textbf{Additional information}: our model introduces paths and concepts into both rule learning and KG embedding while IterE simply depends on triples.

\subsection{Path-Enhanced Models}
In terms of the graph structure of KGs, paths denote the associations between entities apart from relations and are applied to multi-hop reasoning~\cite{Lin:reward-shaping,DeepPath,PathRNN}. PTransE~\cite{PTransE} extends TransE by measuring the similarity between relation and path embeddings. MultiHopKG~\cite{Lin:reward-shaping} explores the answer entities via searching corresponding paths with reinforcement learning. However, these models represent paths in a data-driven fashion, lacking interpretability and accuracy.

\begin{figure*}
  \centering
  \includegraphics[scale=0.38]{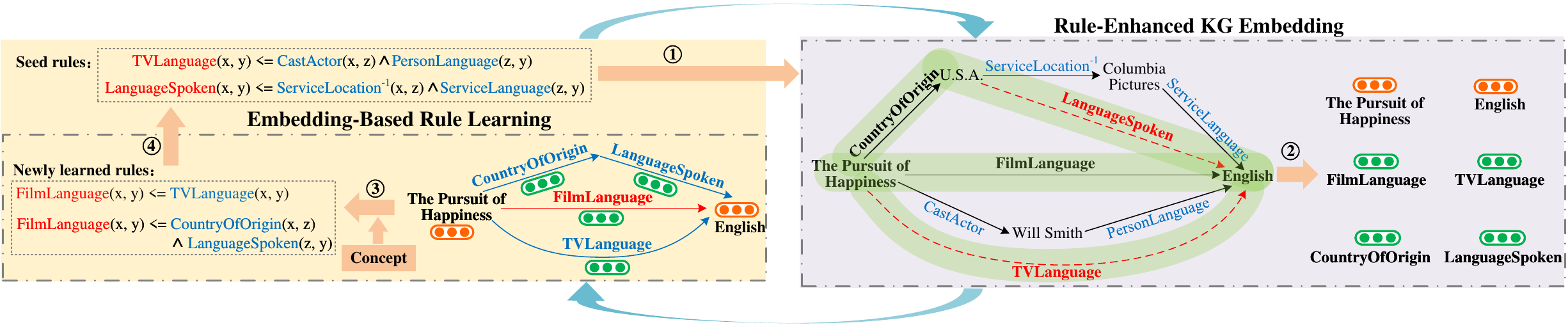}
  \caption{The overall architecture of our developed KG inference model EngineKG in a closed-loop neural-symbolic learning framework. Specific to the rule-enhanced KG embedding module, the green highlighted parts contain the triples and the composed paths via rules, indicating the inputs of the KGE module.}
  \label{fig2}
\end{figure*}

\section{Methodology}
\label{sec:methodology}

In this section, we firstly describe the problem formulation and notation of our work in section \ref{subsec2.1}. Then, following the workflow of EngineKG as shown in Figure \ref{fig2}, we introduce the rule-enhanced KGE module in section \ref{subsec2.2} and the embedding-based rule learning module in section \ref{subsec2.3}.

\subsection{Problem Formulation and Notation}
\label{subsec2.1}

\noindent \textbf{Definition of Closed-Path Rule}. The closed-path (CP) rule or named chain rule is a fragment of the horn rule, which we are interested in for the KGE module and the inference. A CP rule is of the form
\begin{align}
    Head(x,y) \Leftarrow B_1(x,z_1) \wedge B_2(z_1,z_2) \wedge \nonumber \\
     \cdots \wedge B_n(z_{n-1}, y) \label{eq1}
\end{align}
where $B_1(x, z_1)$, $B_2(z_1, z_2)$, $\cdots$, $B_n(z_{n-1}, y)$ denote the atoms in the rule body $Body(x, y)$, and $Head(x, y)$ is the rule head. $B_i$ and $Head$ indicate relations. Standard confidence (SC) and head coverage (HC) are two predefined statistical measurements to assess rules~\cite{Galarrage:AMIE,RLvLR}, which are defined as follows:

\vspace{-3mm}
\begin{small}
\begin{flalign}
    Support &= \#(e, e^{\prime}): Body(e, e^{\prime})\wedge Head(e, e^{\prime}) \label{eq2} \\
    SC &= \frac{Support}{\#(e, e^{\prime}): Body(e, e^{\prime})} \label{eq3} \\
    HC &= \frac{Support}{\#(e, e^{\prime}): Head(e, e^{\prime})} \label{eq4}
\end{flalign}
\end{small}
where $\#(e, e^{\prime})$ indicates the number of entity pairs $(e, e^{\prime})$ that satisfy the condition on the right side of the colon. In general, the rules with $SC$ and $HC$ both higher than 0.7 are regarded as high-quality rules~\cite{IterE}.

\vspace{3pt}
\noindent\textbf{Definition of Path}. A path between an entity pair $(h, t)$ is in the form of [$h \rightarrow r_1 \rightarrow e_1 \rightarrow \cdots \rightarrow r_n \rightarrow t$] where $r_i$ and $e_i$ are the intermediate relation and entity, and the length of a path is the number of the intermediate relations.

\subsection{Rule-Enhanced KGE Module}
\label{subsec2.2}

We aim to learn the entity and relation embeddings from triple facts, rules and paths via neural-symbolic learning. Firstly, we extract the paths via PCRA algorithm~\cite{PTransE}. Apart from other path-finding approaches such as PRA~\cite{Lao:PRA}, PCRA algorithm could measure the reliability of each path for KGE module. Particularly, we develop a joint logic and data-driven path representation mechanism to learn path embeddings.

\vspace{3pt}
\noindent \textbf{Logic-Driven Path Representation} (Intake Stroke).
\label{sec3.21}
The CP rules could compose paths into shorter while more accurate ones for enhancing the representation of paths. For instance, a length-2 path [$The\ Pursuit\ of\ Happiness \xrightarrow{CastActor} Will\ Smith$ $\xrightarrow{PersonLanguage} English$] as shown in Figure \ref{fig2} could be composed into a shorter path (actually a triple) [$The\ Pursuit\ of\ Happiness \xrightarrow{TVLanguage} English$] via the CP rule $ TVLanguage(x, y)$ $\Leftarrow$ $CastActor(x, z)$ $\wedge$ $PersonLanguage(z, y)$. Furthermore, the relation $TVLanguage$ could signify the original multi-hop path.

\vspace{3pt}
\noindent \textbf{Data-Driven Path Representation}. For the scenario that the path cannot be further composed by rules such as the path [$The\ Pursuit\ of\ Happiness \xrightarrow{CountryOfOrigin} U.S.A. \xrightarrow{LanguageSpoken} English$] in Figure \ref{fig2}, we represent this path by summing all the relation embeddings along the path. With the entity pair $(h, t)$ together with the linking path set $\mathcal{P}$, the energy function for measuring the plausibility of the path-specific triple $(h, t, \mathcal{P})$ is designed as
\begin{equation}\small
    E_p(h, t, \mathcal{P}) = \sum_{pi\in \mathcal{P}} {\frac{R(p_i|h, t)}{\sum_{pi\in \mathcal{P}}{R(p_i|h, t)}} \Vert \textbf{h} + \textbf{p}_i - \textbf{t}}\Vert \label{eq5}
\end{equation}
where $\textbf{h}$ and $\textbf{t}$ are the head and tail entity embeddings. $p_i$ denotes the $i$-th path in the path set $\mathcal{P}$ and $\textbf{p}_i$ is the embedding of $p_i$ achieved by the joint logic and data-driven path representation. $R(p_i|h,t)$ indicates the reliability of path $p_i$ between the given entity pair $(h, t)$ obtained by the PCRA algorithm.

\vspace{3pt}
\noindent \textbf{Optimization Objective} (Compression Stroke). Along with the translation-based KGE models, the energy function for formalizing the plausibility of a triple fact (h, r, t) is given as
\begin{equation}
    E_t(h, r, t)= \Vert \textbf{h}+\textbf{r} - \textbf{t} \Vert \label{eq6}
\end{equation}
in which $\textbf{r}$ is the embedding of the relation $r$.

The existing KGE techniques neglect the semantic association between relations. Remarkably, the length-1 rules model the causal correlations between two relations. As shown in Figure \ref{fig2}, the relation pair in the rule $FilmLanguage(x, y) \Leftarrow TVLanguage(x, y)$ should have higher similarity than other relations. Thus, we measure the association between relation pairs as
\begin{equation}
    E_r(r_1, r_2)= \Vert \textbf{r}_1 - \textbf{r}_2 \Vert \label{eq7}
\end{equation}
where $\textbf{r}_1$ and $\textbf{r}_2$ are the embeddings of relations $r_1$ and $r_2$. $E_r(r_1, r_2)$ should be closer to a small value if $r_1$ and $r_2$ appear in a length-1 rule at the same time.

With the energy functions specific to the factual triple, the path representation and the relation correlation, the joint loss function for training is designed as followings:

\vspace{-2mm}
\begin{small}
\begin{align}
     L &=  \sum_{(h,r,t) \in \mathcal{T}} (L_t + \alpha_1 L_p + \alpha_2 L_r) \label{eq8} \\
    L_t &= \sum_{(h^{\prime},r ,t^{\prime}) \in \mathcal{T}^{\prime}}[\gamma_1 + E_t(h,r,t)-E_t(h^{\prime},r ,t^{\prime})]_{+} \label{eq9} \\
    L_p &= \sum_{(h^{\prime},t^{\prime}) \in \mathcal{T}^{\prime}}[\gamma_2 + E_p(h,t,P)-E_t(h^{\prime},t^{\prime}, P^{\prime})]_{+} \label{eq10} \\
    L_r &= \sum_{r_p \in \mathcal{S}}\sum_{r_n \in \mathcal{S}^{\prime}}[\gamma_3 + E_r(r, r_p)-E_r(r, r_n)]_{+} \label{eq11}
\end{align}
\end{small}
where $L$ is the whole training loss consisting of three components: the triple-specific loss $L_t$, the path-specific loss $L_p$, and the relation correlation-specific loss $L_r$. $\alpha_1$ and $\alpha_2$ are the weights of paths and relation correlation, respectively. $\gamma_1$, $\gamma_2$ and $\gamma_3$ are three margins in each loss function. $[x]_{+}$ is the function returning the maximum value between $0$ and $x$. $\mathcal{T}$ is the set of triples observed in the KG and $\mathcal{T}^{\prime}$ is the set of negative samples obtained by random negative sampling. $\mathcal{S}$ is the set of positive relations that are correlated with relation $r$ by length-1 rules and $\mathcal{S}^{\prime}$ is the set of negative relations beyond $\mathcal{S}$ and relation $r$.

We employ mini-batch Stochastic Gradient Descent (SGD) algorithm to optimize the joint loss function for learning entity and relation embeddings. The entity and relation embeddings are initialized randomly and constrained to be unit vectors by the additional regularization term with L2 norm. 

\subsection{Embedding-Based Rule Learning Module}
\label{subsec2.3}


We develop an embedding-based rule learning (Expansion Stroke) to mine high-quality CP rules via conducting the rule searching and the rule pruning efficiently. Remarkably, a path can naturally represent the body of a CP rule. Motivated by this observation, we firstly reuse the paths extracted in section \ref{subsec2.2} and regard these paths as candidate CP rules, which improves the efficiency of rule searching. For instance, given an entity pair $(h, t)$ connected by a relation $r$ and a path [$h \rightarrow r_1 \rightarrow e_1 \rightarrow r_2 \rightarrow e_2 \rightarrow, \cdots \rightarrow e_{n-1} \rightarrow r_n \rightarrow t$], it can be deduced as a CP rule $r(x, y) \Leftarrow r_1(x, z_1) \wedge r_2(z_1, z_2) \wedge \cdots \wedge r_n(z_{n-1}, y)$, where $x$, $y$ and $z_i(i=1,\cdots,n-1)$ are the variables in the rule, and $r_i(i=1,\cdots,n)$ is a relation.

To evaluate the plausibility of candidate CP rules efficiently, we develop a novel rule pruning strategy consisting of two components: \textbf{Embedding-based Semantic Relevance} and \textbf{Concept-based Co-occurrence}. It should be noted that the Concept-based Co-occurrence is available when the KG contains concepts. For the KGs without concepts, employing Embedding-based Semantic Relevance solely is still valid to learn rules.

\vspace{3pt}
\noindent \textbf{Embedding-based Semantic Relevance}.
Intuitively, a candidate rule is plausible if the rule body corresponding to a path $p$ is semantically relevant to the rule head corresponding to the relation $r$. It is noteworthy that we focus on the paths and the CP rules with the lengths no longer than 2 for the trade-off of efficiency and performance. Based on the KG embeddings learned in our KGE module, we could measure the semantic relevance between the body and the head of a candidate rule by the path embedding and relation embedding as well as the score function as
\begin{equation}
    E_{sr}(r, p) = \exp(-\Vert \textbf{r} - \textbf{p} \Vert) \label{eq15}
\end{equation}
where $\textbf{p}$ denotes the embedding of the path $p$.

The embedding-based semantic relevance indicates a global plausibility of a rule from the prospective of relations. Furthermore, a concept-based co-occurrence is proposed to evaluate the local relevance of the arguments in a rule.

\vspace{3pt}
\noindent \textbf{Concept-based Co-occurrences}. 
The neighbor arguments in a high-quality CP rule are expected to share as many same concepts as possible. Given a CP rule $Nationality(x, y) \Leftarrow BornIn(x, z) \wedge LocatedIn(z, y)$, the tail argument of relation $BornIn$ and the head argument of relation $LocatedIn$ should share the concept $Location$. Considering the concepts are far less than the entities, we encode each concept as a one-hot representation to ensure the precision of concept features. The concept embedding of the head or tail argument of an atom can be formalized as
\begin{flalign}
    AC_h(r) &= \frac{1}{\vert(C_h(r)\vert} \sum_{c\in C_h(r)}{OH(c)} \label{eq16} \\
    AC_t(r) &= \frac{1}{\vert(C_t(r)\vert} \sum_{c\in C_t(r)}{OH(c)} \label{eq17}
\end{flalign}
where $AC_h(r)$ and $C_h(r)$ are the concept embedding and concept set in the head argument of an atom containing relation $r$ while $AC_t(r)$ and $C_t(r)$ are that of in the tail argument. $OH(c)$ denotes the one-hot representation of the concept $c$.
 
Specific to a CP rule in the form of $r(x, y) \Leftarrow r_1(x, z_1) \wedge r_2(z_1, z_2) \wedge \cdots \wedge r_n(z_{n-1}, y)$, three types of co-occurrence score functions are designed according to the different positions of the overlapped arguments:
\begin{flalign}
    E_{co}^{h}(r, r_1) &= sim(AC_h(r), AC_h(r_1)) \label{eq18} \\
    E_{co}^{t}(r, r_n) &= sim(AC_t(r), AC_t(r_n)) \label{eq19} \\
    E_{co}^{i}(r_i, r_{i+1}) &= sim(AC_t(r_i), AC_h(r_{i+1}))\label{eq20}
\end{flalign}
where $E_{co}^{h}(r, r_1)$ and $E_{co}^{t}(r, r_n)$ respectively denote the co-occurrence similarities specific to the head arguments and the tail arguments between the rule head and the rule body. $E_{co}^{in}(r_i, r_{i+1})$ represents the co-occurrence similarity between the adjacent arguments in the rule body. $sim(x, y)$ represents the cosine distance function for measuring the similarity between $x$ and $y$.

Then, the whole co-occurrence score function can be achieved by composing all the scores in Eqs. \ref{eq18}-\ref{eq20} as
\begin{equation}
\begin{split}
    E_{co}(r, p) = E_{co}^{h}(r, r_1) + E_{co}^{t}(r, r_n) \\
    + \sum_{i=1}^{n-1}{ E_{co}^{i}(r_i, r_{i+1})} \label{eq21}
\end{split}
\end{equation}
Consequently, the overall score function for evaluating candidate rules is defined as:
\begin{equation}
    E_{cg} = E_{sr}(r, p) + \beta E_{co}(r, p) \label{eq22}
\end{equation}
 where $\beta$ is the weight of the co-occurrence score. We set a threshold and select the candidate rules with the scores calculated by Eq.~\ref{eq22} above the threshold as filtered candidate rules. Afterward, we output the high-quality rules from the filtered candidate rules that satisfy the thresholds of the precise quality criteria namely standard confidence and head coverage defined in Eqs. \ref{eq2}-\ref{eq4}. Then, the updated rule set is obtained via fusing the newly learned rules and the previous rule set (Exhaust Stroke) for the KGE module in the next iteration.

\subsection{Algorithm Flow and Complexity}
\label{sec:algorithm}

It is noteworthy that from the first iteration of EngineKG, our rule learning module could potentially achieve sustainable growth of rules. The entire iteration process will keep running until no fresh rules can be generated. Then, the learned KG embeddings learned in the last iteration are exploited for the KG inference. The detailed algorithm is provided in the appendix of supplementary material.

To evaluate the complexity of our EngineKG model, we denote $n_e$, $n_r$, $n_p$, $n_c$ and $n_t$ as the amount of entities, relations, paths, concepts and triples in a KG. The average length of paths is $l_p$. The embedding dimension of both entities and relations is represented as $d$. The embedding dimension of concepts is $n_c$ due to the one-hot encoding applied for concept representations. Our model complexity of parameter sizes is $\mathcal{O}(n_ed+n_rd+n_{c}^{2})$. For each iteration in training, the time complexity of our model is $\mathcal{O}(n_tn_pl_pd)$.

\begin{table}\scriptsize
 \centering
 \renewcommand\tabcolsep{3.3pt}
 \renewcommand{\arraystretch}{1.0}
 \begin{tabular}{c|ccc|ccc}
 \toprule
Dataset		& \#Relation	& \#Entity	& \#Concept	& \#Train	& \#Valid	& \#Test \\
 \midrule
 FB15K		& 1,345		    & 14,951    & 89            & 483,142	& 50,000	& 59,071 \\
 FB15K237   & 237           & 14,505    & 89            & 272,115   & 17,535    & 20,466 \\
 NELL-995   & 200           & 75,492    & 270           & 123,370   & 15,000    & 15,838 \\
 DBpedia-242 & 298          & 99,744    & 242           & 592,654   & 35,851    & 30,000 \\
 \bottomrule
 \end{tabular}
 \caption{Statistics of the experimental datasets.}
 \label{table1}
 \end{table}

\section{Experiments}
\label{sec:experiment}

\begin{table*}[!t]\scriptsize
\renewcommand{\arraystretch}{1.0}
\centering
\begin{tabular}{c|ccccc|ccccc}
\toprule
\multirow{2}*{Models} & \multicolumn{5}{c|}{FB15K} & \multicolumn{5}{c}{FB15K237} \\
	& MR	& MRR	& Hits@10	& Hits@3  & Hits@1	& MR	& MRR	& Hits@10	& Hits@3  & Hits@1\\
\midrule
TransE~\cite{Bordes:TransE}         & 117   & 0.534     & 0.775     & 0.646     & 0.386       & 228     & 0.289     & 0.478     & 0.326     & 0.193 \\
ComplEx~\cite{Trouillon:ComplEx}    & 197   & 0.346     & 0.593     & 0.405     & 0.221       & 507     & 0.236     & 0.406     & 0.263     & 0.150 \\
RotatE~\cite{RotatE}    & \underline{39}    & 0.612     & 0.816     & 0.698     & 0.488       & \underline{168}   & 0.317     & 0.553     & 0.375     & 0.231 \\
QuatE~\cite{QuatE}                  & 40    & 0.765     & 0.878     & 0.819     & 0.693       & 173     & 0.312     & 0.495     & 0.344     & 0.222 \\
HAKE~\cite{HAKE}                    & 42    & 0.678     & 0.839     & 0.761     & 0.570       & 183     & 0.344     & 0.542     & 0.382     & 0.246 \\
DualE~\cite{DualE}                    & 43    & 0.759     & 0.882     & 0.820     & 0.681       & 202     & 0.332     & 0.522     & 0.367     & 0.238 \\
\midrule
MultiHopKG~\cite{Lin:reward-shaping}              & -    & 0.670     & 0.769     & 0.708     & 0.612       & -     & 0.385     & 0.562     & 0.429     & 0.298 \\
RNNLogic~\cite{RNNLogic}              & 244    & 0.496     & 0.669     & 0.544     & 0.405       & 620     & 0.280     & 0.428     & 0.306     & 0.205 \\
RPJE~\cite{RPJE}                    & 40    & \underline{0.811}     & \underline{0.898}     & \underline{0.832}     & \underline{0.762}     & 207     & \underline{0.443}     & \underline{0.579}         & \underline{0.479}      & \underline{0.374}  \\
IterE~\cite{IterE}                  & 85    & 0.577     & 0.807     & 0.663     & 0.451       & 463     & 0.210     & 0.355     & 0.227     & 0.139 \\
\midrule
\textbf{EngineKG~(Ours)}	        & \textbf{20}   & \textbf{0.854}    & \textbf{0.933}    & \textbf{0.885}	& \textbf{0.810}     & \textbf{121} & \textbf{0.555}  & \textbf{0.707}      & \textbf{0.590}   & \textbf{0.479}  \\
\bottomrule
\toprule
\multirow{2}*{Models} & \multicolumn{5}{c|}{DBpedia-242} & \multicolumn{5}{c}{NELL-995} \\
	& MR	& MRR	& Hits@10	& Hits@3  & Hits@1	& MR	& MRR	& Hits@10	& Hits@3  & Hits@1\\
\midrule
TransE~\cite{Bordes:TransE}         & 1996   & 0.256     & 0.539    & 0.395     & 0.075    & 8650      & 0.167     & 0.354     & 0.219      & 0.068 \\
ComplEx~\cite{Trouillon:ComplEx}    & 3839   & 0.196     & 0.387    & 0.230     & 0.104    & 11772     & 0.169     & 0.298     & 0.185      & 0.106 \\
RotatE~\cite{RotatE}        & \underline{1323}      & 0.308     & 0.594     & 0.422     & 0.143        & 9620       & 0.292     & 0.444    & 0.325  & 0.216 \\
QuatE~\cite{QuatE}        & 1618    & 0.411     & 0.612         & 0.491     & 0.293        & 12296      & 0.281    & 0.422     & 0.315  & 0.207 \\
HAKE~\cite{HAKE}      & 1522     & 0.379     & 0.551     & 0.432    & 0.283     & 13211    & 0.245     & 0.370      & 0.283   & 0.175 \\
DualE~\cite{DualE}     & 1363    & 0.360     & 0.592     & 0.439     & 0.232       & 11529     & 0.292     & 0.447     & 0.329     & 0.214 \\
\midrule
MultiHopKG~\cite{Lin:reward-shaping}              & -    & 0.520     & \underline{0.625}     & 0.530     & 0.426       & -     & \underline{0.416}     & 0.474     & 0.345     & 0.275 \\
RNNLogic~\cite{RNNLogic}              & 7857    & 0.344     & 0.514     & 0.390     & 0.253       & 15772     & 0.335     & 0.422     & 0.356     & \underline{0.290} \\
RPJE~\cite{RPJE}        & 1770      & \underline{0.521}     & 0.576  & \underline{0.542}     & \underline{0.487}    & \underline{6291}     & 0.360     & \underline{0.496}  & \underline{0.401}     & 0.288  \\
IterE~\cite{IterE}      & 5016      & 0.190     & 0.326     & 0.215     & 0.120     & 12998     & 0.233     & 0.327     & 0.246     & 0.185 \\
\midrule
\textbf{EngineKG~(Ours)}    & \textbf{1275}     & \textbf{0.523}    & \textbf{0.647}    & \textbf{0.551}	& \textbf{0.501}    & \textbf{5243}      & \textbf{0.454}     & \textbf{0.506}      & \textbf{0.407} 	& \textbf{0.293}  \\
\bottomrule
\end{tabular}
\caption{Link prediction results on four datasets. \textbf{Bold} numbers are the best results, and the second best is \underline{underlined}.}
\label{table2}
\end{table*}

\subsection{Experimental Setup}

\noindent \textbf{Datasets}.
Four datasets containing ontological concepts are employed for our experiments, including FB15K~\cite{Bordes:TransE}, FB15K237~\cite{FB15k237}, NELL-995 and DBpedia-242. Particularly, NELL-995 here is a re-split of the original dataset~\cite{DeepPath} into training/validation/test sets. DBpedia-242 is generated from the commonly-used KG DBpedia~\cite{Lehmann:dbpedia} to ensure each entity in the dataset has a concept. The statistics of the experimental datasets are listed in Table~\ref{table1}.


\vspace{3pt}
\noindent \textbf{Baselines}.
We compare our model EngineKG with two categories of baselines: 

(1) The traditional KGE models depending on triple facts: TransE~\cite{Bordes:TransE}, ComplEx~\cite{Trouillon:ComplEx}, RotatE~\cite{RotatE}, QuatE~\cite{QuatE}, HAKE~\cite{HAKE} and DualE~\cite{DualE}.

(2) The models using paths or rules: the path-based model MultiHopKG~\cite{Lin:reward-shaping}, the rule learning-based models RNNLogic~\cite{RNNLogic} and RPJE~\cite{RPJE}, and the model combining rules with KG embeddings IterE~\cite{IterE}. The evaluation results of these baselines are obtained by employing their open-source codes with the suggested hyper-parameters.

\vspace{3pt}
\noindent \textbf{Training Details}.
We implement our model in C++ and on Intel i9-9900 CPU with memory of 64G. For the fair comparison, the embedding dimension of all the models is fixed as 100, the batch size is set to 1024 and the number of negative samples is set to 10. Specific to our model, during each iteration, the maximum training epoch is set to 1000, the standard confidence and the head coverage are selected as 0.7 and 0.1 for better performance. The entity and relation embeddings are initialized randomly. We employ grid search for selecting the best hyper-parameters on the validation dataset.

\vspace{3pt}
\noindent \textbf{Evaluation Metrics}.
\label{sec:evaluation}
Take head entity prediction for instance, we fill the missing head entity with each entity $e$ in the KG, and score a candidate triple $(e, r, t)$ according to the following energy function together with the path information:
\begin{equation}
    E_e(e, r, t, \mathcal{P}) = E_t(e, r, t) + \alpha_1 E_p(e, t, \mathcal{P}) \label{eq23}
\end{equation}
in which we reuse the energy functions in Eq. \ref{eq5} and Eq. \ref{eq6}, and $\mathcal{P}$ is the path set consisting of all the paths between entities $e$ and $t$. We rank the scores of the candidate triples in ascending order. Tail entity prediction is in the similar way.

\begin{table*}\small
  \centering
  \renewcommand\tabcolsep{10.0pt}
  \renewcommand{\arraystretch}{1.2}
  \begin{tabular}{c|cccc|cccc}
  \toprule
  \multirow{2}*{FB15K}	& \multicolumn{4}{c|}{Head Entities Prediction}	& \multicolumn{4}{c}{Tail Entities Prediction}\\
    & 1-1	& 1-N	& N-1	& N-N	& 1-1	& 1-N	& N-1	& N-N\\
  \hline
  TransE~\cite{Bordes:TransE}	& 0.356	& 0.626	& 0.172	& 0.375	& 0.349	& 0.146	& 0.683	& 0.413\\
  RotatE~\cite{RotatE}	& 0.895	& 0.966	& 0.602	& 0.893	& 0.881 & 0.613	& 0.956	& 0.922 \\
  HAKE~\cite{RotatE}	    & 0.926	& 0.962	& 0.174	& 0.289	& \underline{0.920} & 0.682	& \textbf{0.965}	& 0.805 \\
  DualE~\cite{DualE}	    & 0.912	& \underline{0.967}	& 0.557	& 0.901	& 0.915 & 0.662	& 0.954	& 0.926 \\
  \hline
  MultiHopKG~\cite{Lin:reward-shaping}	& -	& -	& -	& -	& 0.893	& 0.576	& 0.921	& 0.763\\
  RPJE~\cite{RPJE}	& \underline{0.942}	& 0.965	& \underline{0.704}	& \underline{0.916}	& \textbf{0.941}	& \underline{0.839}	& 0.953	& \underline{0.933}\\
  \hline
  \textbf{EngineKG}	& \textbf{0.943}	& \textbf{0.968}	& \textbf{0.835}	& \textbf{0.929}	& \textbf{0.941}	& \textbf{0.904}	& \underline{0.957}	& \textbf{0.949}\\
  \hline
  \hline
  \multirow{2}*{FB15K237}	& \multicolumn{4}{c|}{Head Entities Prediction}	& \multicolumn{4}{c}{Tail Entities Prediction}\\
    & 1-1	& 1-N	& N-1	& N-N	& 1-1	& 1-N	& N-1	& N-N\\
  \hline
  TransE~\cite{Bordes:TransE}	& 0.356	& 0.626	& 0.172	& 0.375	& 0.349	& 0.146	& 0.683	& 0.413\\
  RotatE~\cite{RotatE}	& 0.547	& 0.672	& \underline{0.186}	& 0.474	& 0.578	& 0.140	& 0.876	& 0.609 \\
  HAKE\cite{HAKE}	    & \underline{0.791}	& \textbf{0.833}	& 0.098	& 0.237	& \underline{0.794} & \underline{0.372}	& \textbf{0.938}	& \textbf{0.803} \\
  DualE\cite{DualE}	    & 0.516	& 0.637	& 0.153	& 0.471	& 0.526 & 0.135	& 0.860	& 0.607 \\
  \hline
  MultiHopKG\cite{Lin:reward-shaping}	& -	& -	& -	& -	& 0.417	& 0.026	& 0.794	& 0.457\\
  RPJE\cite{RPJE}	& 0.692	& 0.476	& 0.180	& \underline{0.575}	& 0.669	& 0.197	& 0.691	& 0.685 \\
  \hline
  \textbf{EngineKG}	& \textbf{0.792}	& \underline{0.743}	& \textbf{0.629}	& \textbf{0.651}	& \textbf{0.807}	& \textbf{0.399}	& \underline{0.881}	& \underline{0.757} \\
  \bottomrule
  \end{tabular}
  \caption{Link prediction results on FB15K and FB15K237 on various relation properties (Hits@10). MultiHopKG could only predict tail entities rather than head entities.}
  \label{table3}
  \end{table*}
  
We employ three frequently-used metrics: (1) Mean rank (MR) and (2) Mean reciprocal rank (MRR) of the triples containing the correct entities. (3) Hits@n is the proportion of the correct triples ranked in the top n. The lower MR, the higher MRR and the higher Hits@n declare the better performance. All the results are ``filtered'' by wiping out the candidate triples that are already in the KG~\cite{Wang:TransH}.

\subsection{Results of Link Prediction}
\label{linkprediction}

The evaluation results of link prediction are reported in Table \ref{table2}. \textbf{Firstly}, our model EngineKG significantly and consistently outperforms all the state-of-the-art baselines on all the datasets and all the metrics. Compared to best-performing models RotatE and RPJE on MR, EngineKG achieves performance gains of 95.0\%/42.4\%/3.7\%/83.5\%  compared to RotatE and 100.0\%/71.1\%/38.8\%/20.0\% against RPJE on datasets FB15K/FB15K237/DBpedia-242/NELL-995. Particularly, on FB15K and FB15K237, the difference between the best performing baseline RPJE and our developed model is statistically significant under the paired at the 99\% significance level. \textbf{Secondly}, our model achieves better performance than the traditional models that utilize triples alone, indicating that EngineKG is capable of taking advantage of extra knowledge including rules and paths as well as concepts, which all benefit to improving the performance of the whole model. \textbf{Thirdly}, EngineKG further beats IterE, illustrating the superiority of exploiting rules and paths for KG inference in a joint logic and data-driven fashion.

\subsection{Evaluation on Various Relation Properties}

The relations can be classified into four categories: One-to-One (1-1), One-to-Many (1-N), Many-to-One (N-1), and Many-to-Many (N-N). We select some well-performing models observed in Table \ref{table2} as the baselines in this section. Table \ref{table3} exhibits that EngineKG achieves more performance gains on complex relations compared to other baselines. More interestingly, specific to the most challenging tasks (highlighted) namely predicting head entities on N-1 relations and tail entities on 1-N relations, our model consistently and significantly outperforms the outstanding baselines RotatE and RPJE by achieving the performance improvements: 38.7\%/47.5\% on FB15K and 238.2\%/185.0\% on FB15K237 compared to HAKE while 18.6\%/7.75\% on FB15K and 249.4\%/102.5\% on FB15K237 compared to RPJE. These results all demonstrate that the paths and the generated rules enrich the associations between entities and relations, contributing to better performance of KG inference on complex relations.

\subsection{Performance Evaluation Over Iterations}

\begin{figure}
  \centering
  \includegraphics[scale=0.4]{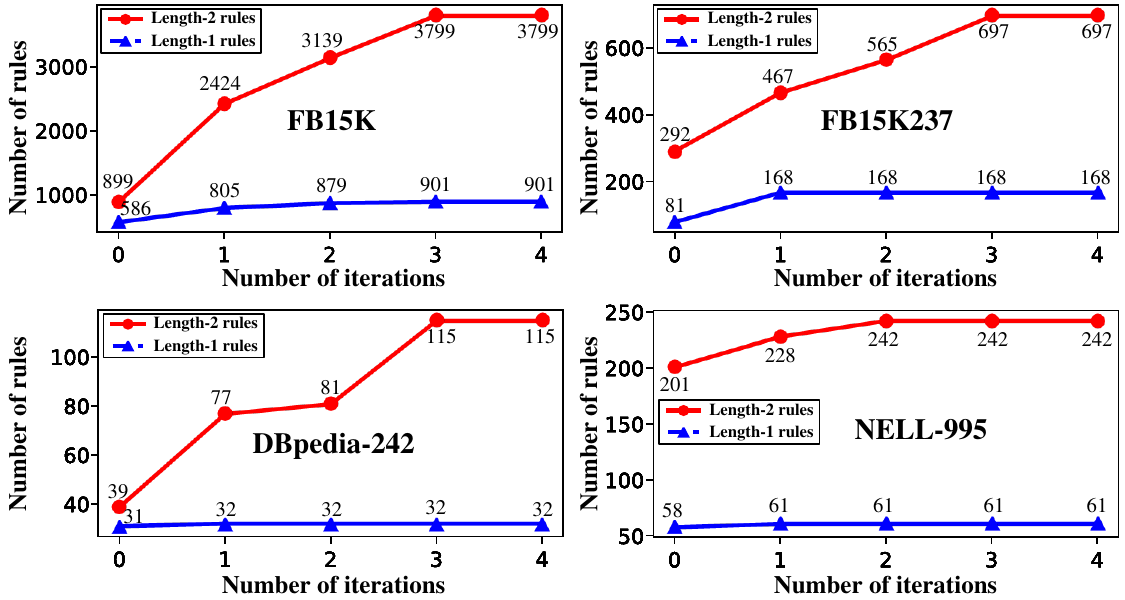}

  \caption{The number of rules over iterations.}
  \label{fig:iterative}
\end{figure}

We evaluate the performance of our rule learning module on the learning time and the amount of rules compared to the excellent rule learning tool AMIE+. For generating high-quality rules, our model takes 6.29s/2.26s/1.55s/10.50s in an iteration on average while AMIE+ takes 79.19s/26.83s/5.35s/105.53s on datasets FB15K/FB15K237/NELL-995/DBpedia, illustrating the higher efficiency of EngineKG. In Figure \ref{fig:iterative}, the amount of rules mined by AMIE+ is shown as that at the initial iteration. Thus, we can discover that the quantity of rules generated in the first iteration and the third iteration is twice and three times the number of rules obtained by AMIE+.

\begin{figure}
  \centering
  \includegraphics[scale=0.42]{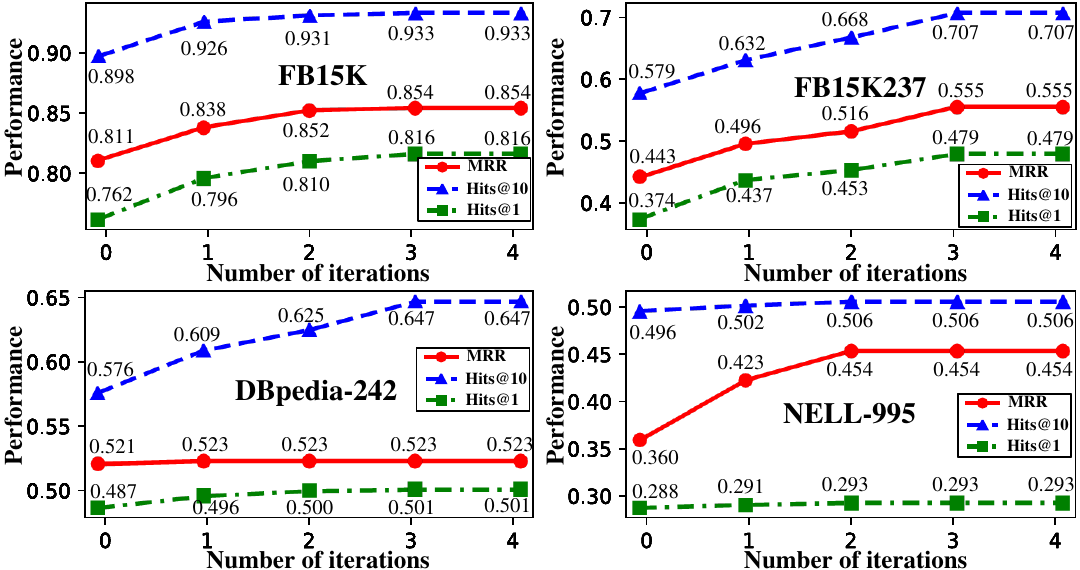}
  \caption{The performance curves of MRR, Hits@10 and Hits@1 over iterations on four datasets.}
  \label{fig:perfocurve}
\end{figure}

More specifically, Figure \ref{fig:iterative} exhibits the number of rules and Figure \ref{fig:perfocurve} indicates the performance curves on the four datasets over iterations. Notably, the number of rules and the performance continue to grow as the iteration goes on and converge after three iterations on all the datasets. These results illustrate that: (1) Rule learning and KGE modules in our model indeed complement each other and benefit in not only producing more high-quality rules but also obtaining better inference results. (2) More rules are effective to improve the performance of KG inference. (3) The iteration process will gradually converge along with the rule learning.

\subsection{Ablation Study}

\begin{figure}
  \centering
  \includegraphics[scale=0.55]{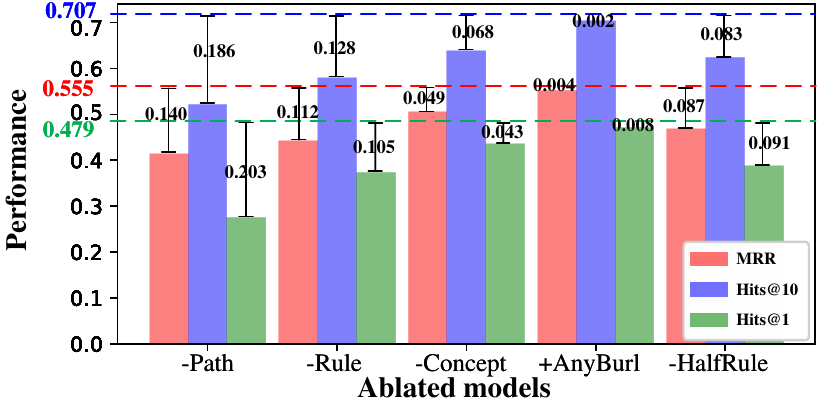}
  \caption{Ablation study on FB15K237. The dash lines indicate the performance of our whole model on MRR (red), Hits@10 (blue) and Hits@1 (green).}
  \label{fig:ablation}
\end{figure}

\begin{figure}
  \centering
  \includegraphics[scale=0.77]{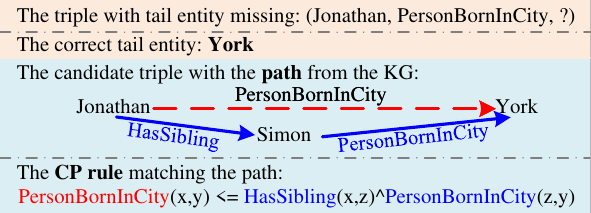}
  \caption{An example of the interpretable tail entity prediction via path and rule on NELL-995.}
  \label{fig:casestudy}
\end{figure}

To evaluate each contribution in our whole model EngineKG, we observe the performance on FB15K237 as to the five different ablated settings: (1) Omitting paths (-Path). (2) Omitting rules (-Rule). (3) Omitting concepts (-Concept) by removing concept-based co-occurrences in the rule learning. (4) Replacing rule mining tool AMIE+ with AnyBurl~\cite{AnyBurl} for obtaining the seed rules (+AnyBurl). (5) Employing only half of the seed rules and without iteration (-HalfRule). Figure \ref{fig:ablation} shows that the performance of our whole model is better than that of all the ablated models except for ``+AnyBurl'', demonstrating that all the components in our designed model are effective and our model is free of any rule mining tool for obtaining the seed rules. Besides, removing paths and rules both have significant impacts on the performance, which suggests the paths and rules in our model play more vital role in KG inference.

\subsection{Case Study}

As shown in Figure \ref{fig:casestudy}, although the head entity $Jonathan$ \normalsize and the candidate tail entity $York$ \normalsize are not linked by any direct relation in the KG, there is an explicit path between them. This path can be represented as the relation $PersonBornInCity$ \normalsize deduced by a matched CP rule. The path and the rule together boost the score of the correct candidate entity $York$ calculated by Eq. \ref{eq23}, and especially provide the interpretability of the result.

\section{Conclusion and Future Work}
\label{sec:conclusion}

In this paper, we develop a novel closed-loop neural-symbolic learning framework EngineKG for KG inference by jointly rule learning and KGE while exploiting paths and concepts. In the KGE module, both rules and paths are introduced to enhance the semantic associations and interpretability for learning the entity and relation embeddings. In the rule learning module, paths and KG embeddings together with entity concepts are leveraged in the designed rule pruning strategy to generate high-quality rules efficiently and effectively. Extensive experimental results on four datasets illustrate the superiority and effectiveness of our approach compared to some state-of-the-art baselines. In the future, we will investigate combining other semantics such as contextual descriptions of entities, and attempt to apply our model in dynamic KGs.

\bibliography{acl_latex}
\bibliographystyle{acl_natbib}

\end{document}